\documentclass{esannV2}
\usepackage[dvips]{graphicx}
\usepackage[latin1]{inputenc}
\usepackage{amssymb,amsmath,array}
\usepackage{graphicx}
\usepackage{bm, bbm}
\usepackage{nicefrac}
\usepackage{xspace}
\usepackage{xcolor}
\usepackage{subcaption}

\newcommand{\modelname}{GenKSC\xspace}
\usepackage{algorithm}
\usepackage{algorithmic}

\begin{document}
\title{Generative Kernel Spectral Clustering }

\author{David Winant\thanks{Equal contribution.} ,  
Sonny Achten\footnotemark[1] \thanks{Corresponding author: \texttt{sonny.achten@kuleuven.be}} ,
Johan A. K. Suykens
\thanks{\textbf{This work has been accepted for publication at ESANN 2025.} The research leading to these results has received funding from the European Research Council under the European Union's Horizon 2020 research and innovation program / ERC Advanced Grant E-DUALITY (787960). This paper reflects only the authors' views and the Union is not liable for any use that may be made of the contained information. This research received funding from the Flemish Government (AI Research Program); iBOF/23/064; KU Leuven C1 project C14/24/103. Johan Suykens, David Winant, and Sonny Achten are also affiliated to Leuven.AI - KU Leuven institute for AI, B-3000, Leuven, Belgium.}
%
\vspace{.3cm}\\
%
ESAT-Stadius, KU Leuven \\
Kasteelpark Arenberg 10, B-3001 Leuven, Belgium
%
\vspace{.1cm}\\
}

\maketitle

\begin{abstract}
Modern clustering approaches often trade interpretability for performance, particularly in deep learning-based methods. We present Generative Kernel Spectral Clustering (\modelname), a novel model combining kernel spectral clustering with generative modeling to produce both well-defined clusters and interpretable representations. By augmenting weighted variance maximization with reconstruction and clustering losses, our model creates an explorable latent space where cluster characteristics can be visualized through traversals along cluster directions. Results on MNIST and FashionMNIST datasets demonstrate the model's ability to learn meaningful cluster representations.
\end{abstract}

\section{Introduction}
Clustering is a key technique in data analysis, used to uncover patterns in unlabeled data by grouping similar instances. While modern neural network-based clustering methods often achieve impressive performance, they frequently lack interpretability, making it difficult to understand the characteristics that define each cluster. This limitation is especially concerning in sensitive domains---such as healthcare, finance, and security---where understanding the basis of clustering results is critical for transparency, trust, and informed decision-making.

Deep clustering methods often lack interpretability \cite{10585323}, while interpretable methods rarely use deep architectures and rely on post hoc explanations, limiting their ability to capture complex patterns \cite{hu2024interpretableclusteringsurvey}. This gap in the literature shows a need for models that combine the representational power of deep learning with the transparency of interpretable clustering. To address this challenge, we propose \modelname, a novel interpretable clustering model that combines clustering with generative modeling. \modelname produces well-defined clusters while allowing users to interpret the distinguishing features of each group. Our approach leverages the latent structure of a kernel spectral clustering (KSC) framework, integrating it with a generative restricted kernel machine. Augmented loss terms further guide the model to form clear, interpretable clusters, ensuring the learned representations are both accurate and explorable. By bridging the gap between clustering and interpretability, \modelname advances explainable AI, offering a valuable tool for applications where \emph{understanding} the clustering result is critical.

\section{Preliminaries and Related Work}

Restricted Kernel Machines (RKM) \cite{suykens2017deep}, introduced conjugate feature duality in a kernel-based setting, facilitating both supervised and unsupervised learning and supporting deep kernel learning. The Stiefel-RKM \cite{pandey2022Stiefel} is a generative model that achieves interpretability through a disentangled latent space within a kernel principal component analysis framework. 

Exploring latent spaces has been enabled in other generative models, such as variational autoencoders \cite{kingma2013auto}, as well as the combination with a clustering model in ClusterGAN \cite{Mukherjee_Asnani_Lin_Kannan_2019} where an adversarial loss was used along with a latent space clustering objective to preserve clustering. However, ClusterGAN does not inherently provide interpretability of the learned clusters. 

\begin{figure}[h!]
    \centering
    \includegraphics[scale=0.2]{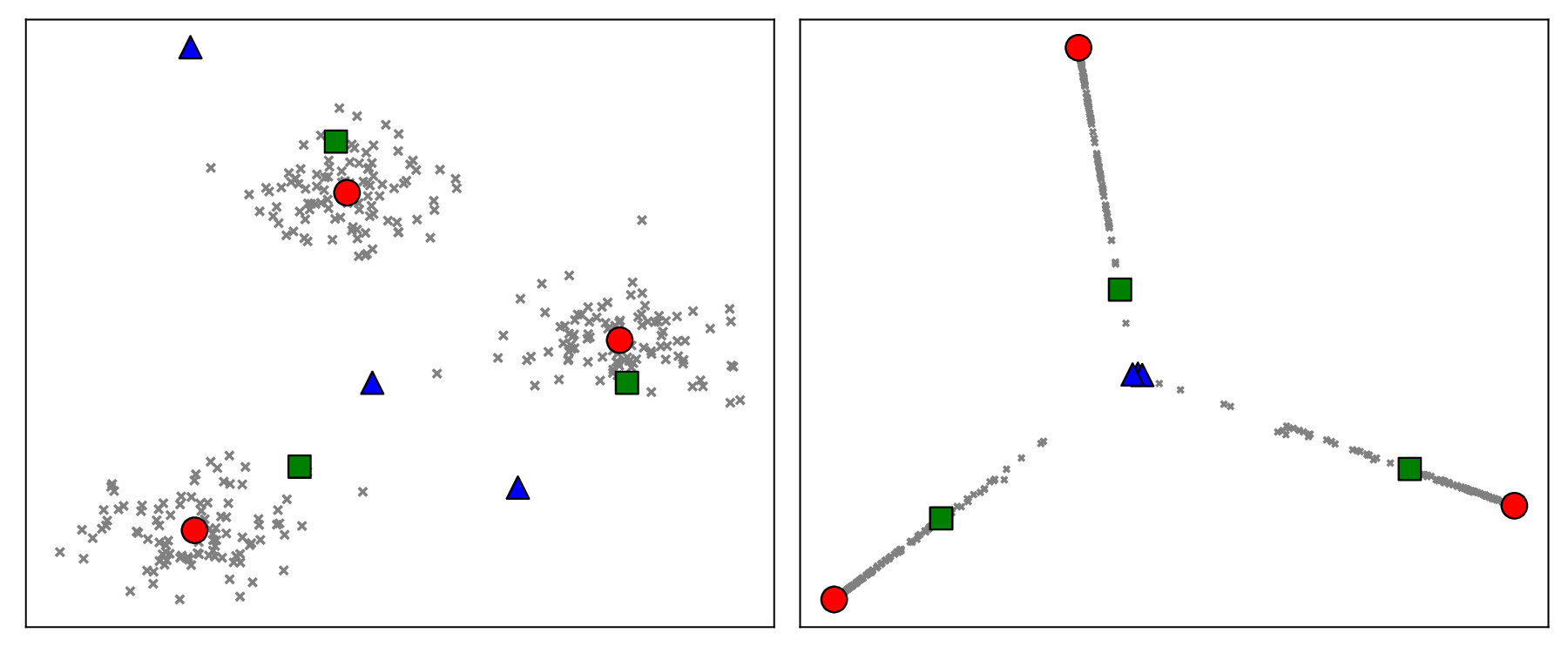} 
    \caption{Visualization of latent structure with KSC \cite{alzate2008multiway}. \textbf{Left}: The data in its original space. \textbf{Right}: The spectral embeddings in the eigenspace of the first two components. Observe that the cluster prototypes align at the tips of the lines. A radial basis function kernel is used in this example.}
    \label{fig:KSC_structure}
\end{figure}

Alzate and Suykens \cite{alzate2008multiway} introduced kernel spectral clustering (KSC)---a nonlinear extension to spectral clustering---by framing it as a weighted principal component analysis (PCA) problem in a (implicit) feature space. 
The solution to KSC is formulated as an eigendecomposition problem:
\begin{equation}\label{eq:KSC}
\bm{D}^{-1}\bm{K}\bm{H}=\bm{H}\bm{\Lambda},
\end{equation}
where $\bm{K}$ represents the kernel matrix with $K_{ij}=\left<\phi(\bm{x}_i),\phi(\bm{x}_j)\right>_\mathcal{H}$, and $\mathcal{H}$ denotes the reproducing kernel Hilbert space associated with the kernel. 
Here, $\bm{D}$ is the diagonal degree matrix with elements $D_{ii}=\sum_j K_{ij}$, $\bm{H}=[\bm{h}_1, \dots, \bm{h}_n]^\top$ represents the spectral embeddings, and $\bm{\Lambda}$ is a diagonal matrix containing the corresponding eigenvalues along its diagonal. As illustrated in Fig. \ref{fig:KSC_structure}, distinct linear structures emerge in the $(k-1)$ dimensional eigenspace spanned by the highest principal components when data are clustered into $k$ groups. Given this distinct line structure, cosine similarity is highly suitable for both cluster assignments and cluster quality evaluation \cite{langone2013soft}.

\section{The Generative Kernel Spectral Clustering Model}
The proposed model leverages a weighted variance maximization framework, which shares foundational connections with the kernel spectral clustering model \eqref{eq:KSC}. We use a parametric feature map that is trained simultaneously with the spectral clustering problem, rather than relying on a predefined kernel function. To enable both clustering and generative tasks, the model incorporates a reconstruction loss along with an unsupervised clustering loss.

\subsection{Spectral Clustering Loss}

For a dataset $\{\bm{x}_i\}_{i=1}^n$, $\bm x_i \in \mathbb{R}^d$, and feature map $\phi: \mathcal{I} \subseteq \mathbb{R}^d \mapsto \mathcal{F} \subseteq \mathbb{R}^{d_f}$, the weighted variance maximization problem in $\mathcal{F}$ can be formulated as:   
\begin{equation}\label{eq:weigthed_varmax}
    \max_{\bm{U}}  \frac{1}{2}\sum^n_{i=1}D_{ii}^{-1}\left\|\bm{U}^{\top}\phi(\bm{x}_i)\right\|_2^2 \qquad \text{s.t. } \bm{U}^\top \bm{U} = \bm{I}_s, 
\end{equation} 
where the weighting scalars are the inverse degrees of the kernel matrix, $D_{ii} = \sum_j \phi(\bm{x}_i)^{\top}\phi(\bm{x}_j)$, and $\bm U \in \mathbb{R}^{d_f \times s}$ is a projection matrix with $s<d_f$.\footnote{We assume that the feature map is centered with respect to the weighting scheme, which can be implemented by updating $\phi(\bm{x}) \xleftarrow{} \phi(\bm{x}) - \sum_i D_{ii}^{-1}\phi(\bm{x}_i)/\sum_i{D_{ii}^{-1}}$.} The optimal $\bm{U}^*$ spans the eigenspace of the top $s$ components of this weighted PCA problem. In traditional KSC problems, $s$ is set to $k-1$, where $k$ is the number of clusters to infer. In our model, we allow $s > k-1$, creating a richer latent representation for generation. The stationarity conditions of the optimization problem in \eqref{eq:weigthed_varmax} demonstrate the equivalence between the problems \eqref{eq:KSC} and \eqref{eq:weigthed_varmax}. The kernel is defined as $K_{ij} = \phi(\bm{x}_i)^\top\phi(\bm{x}_j)$, and the principal component score vectors $\bm{e}_i=\bm{U}^{\top}\phi(\bm{x}_i)$ relate to the spectral embeddings $\bm{h}_i$ as: $\bm{e}_i = D_{ii}\bm{h}_i\bm{\Lambda}$.\footnote{Refer to \cite{achten2023duality} for a detailed mathematical comparison between equivalent primal and dual formulations within a PCA framework.}

\subsection{Augmented Losses}
Problem \eqref{eq:weigthed_varmax} formulates the KSC problem for a given feature map. A novel aspect of this work is the use of learnable feature mappings, such as neural networks, in the KSC framework. This requires the addition of augmented loss terms. Following Pandey et al. \cite{pandey2022Stiefel}, we incorporate an inverse mapping $\psi:\mathcal{F}\mapsto\mathcal{I}$, enabling an encoder-decoder architecture that facilitates representation learning. We denote the parametric feature map and its approximate inverse as $\phi(\cdot;\bm\theta_\phi)$ and $\psi(\cdot;\bm\theta_\psi)$, respectively. To optimize the parameters $\bm\theta_\phi$ and $\bm\theta_\psi$, we introduce a reconstruction error term. Since feature representations are projected onto the eigenspace through $\bm{U}$, this reconstruction depends on the KSC problem \eqref{eq:genKSC_loss}:
\begin{equation*}
    \mathcal{L}_{\rm rec} = \sum_{i=1}^n \left\| \bm x_i - \psi\left(\bm{UU}^\top\phi(\bm{x}_i;\bm\theta_\phi) ;\bm\theta_\psi\right)\right\|^2_2.
\end{equation*}

Additionally, we introduce a cluster loss term. As depicted in Fig. \ref{fig:KSC_structure}, effective clustering in the KSC framework produces a line-structured distribution in the score vector space. We predefine $k$ directions for these lines using cluster codes ($\{\bm s_c\}_{c=1}^k$) and minimize the cosine distance of each representation to its closest cluster code. We set these cluster codes as the vertices of a regular $(k-1)$-simplex ensuring maximal angular separation between the equidistant vertices, leading to clear and distinct cluster directions. The cosine distance of a point to a cluster code, along with the total cluster loss, is given by: 
\begin{equation*}\label{eq:cosdistance}
    d^{\rm cos}_{ic}  =  1 - \frac{\bm{e}_i^{\prime\top}\bm{s}_c}{\left\|\bm{e}_i^{\prime}\right\|_2\left\|\bm{s}_c\right\|_2} \qquad \mathcal{L}_{\rm cl} = \sum_{i=1}^n \min_c d^{\rm cos}_{ic},
\end{equation*}
where $\bm e_i^\prime = \bm e_{i,1:k-1}$ represents the first $(k-1)$ elements of the score vectors. Note that the solution of \eqref{eq:weigthed_varmax} yields an arbitrary rotation of the first $s$ components, distributing cluster information across all $s$ components rather than solely within the first $(k-1)$. The proposed cosine distance loss encourages optimal rotation in the first $(k-1)$ components, and further enhances linearity within this subspace.

\subsection{The \modelname Model}
Combining the above loss terms and adding regularization on feature representations, we arrive at the optimization problem for \modelname:
\begin{multline}\label{eq:genKSC_loss}
    \min_{\bm{U}, \bm\theta_\phi, \bm\theta_\psi} \sum^n_{i=1} \left( - \frac{1}{2}D_{ii}^{-1}\left\|\bm{U}^{\top}\phi(\bm{x}_i;\bm\theta_\phi)\right\|_2^2 + \left\|\phi(\bm x_i;\bm\theta_\phi)\right\|^2_2 \right.\\ 
    \left. + \ \eta_{\rm rec} \left\| \bm x_i - \psi\left(\bm{UU}^\top\phi(\bm{x}_i;\bm\theta_\phi) ;\bm\theta_\psi\right)\right\|^2_2 + \eta_{\rm cl} \min_c d^{\rm cos}_{ic} \right) \qquad  \text{s.t. } \bm{U}^\top \bm{U} = \bm{I}_s,
\end{multline}
where $\eta_{\rm rec}$ and $\eta_{\rm cl}$ are hyperparameters balancing the contributions of the respective loss terms; and where feature map parameters $\bm\theta_\phi$, inverse feature map parameters $\bm\theta_\psi$, and projection matrix $\bm U$ are the training parameters.  

This formulation effectively constructs a spectral clustering problem within a feature space, while simultaneously learning the feature representations.  After training, a new point $\bm{e}^*$ in the score variable space can be selected by targeting a specific cluster center to generate a representative datapoint, or sampled randomly to explore the latent space. The corresponding datapoint is then computed as $\bm{x}^* = \psi(\bm{U} \bm{e}^*)$.

\section{Experiments}

\subsection{Datasets and Model Details}
For clarity, we select a subset of the MNIST dataset, containing only the first three digit classes (0, 1, and 2), for a total of 18,732 images; termed MNIST012. 
For a more challenging experiment, we use the FashionMNIST dataset. For both datasets, convolutional neural networks were selected as parametric feature maps $\phi(\cdot;\bm\theta_\phi)$ with the approximate inverse feature map $\psi(\cdot;\bm\theta_\psi)$ using a mirrored architecture with transposed convolutions. For MNIST012, latent space dimensions were set to $s=10$ with $k=3$, using three convolutional and two linear layers in the encoder. For FashionMNIST, we used $s=40$, $k=10$, with similar architectures as in ClusterGAN. To avoid clustering on arbitrary features, cluster loss was excluded from the objective function for the first 10 epochs on MNIST012 and 32 epochs on FashionMNIST, allowing the model to develop meaningful representations before clustering. We use Cayley ADAM \cite{li2019} to enforce the orthonormal constraint on $\bm{U}$. Loss weights $\eta_{\rm rec}$ and $\eta_{\rm cl}$ were set to 1 for MNIST012, while for FashionMNIST, the values $\eta_{\rm rec} = 0.001$ and $\eta_{\rm cl} = 0.008$ were determined through hyperparameter tuning based on the average membership strength criterion, as in \cite{langone2013soft}.

\subsection{Results}
\begin{figure}[h!]
    \centering
    \begin{subfigure}{0.35\textwidth}
        \centering
        \includegraphics[width=\linewidth]{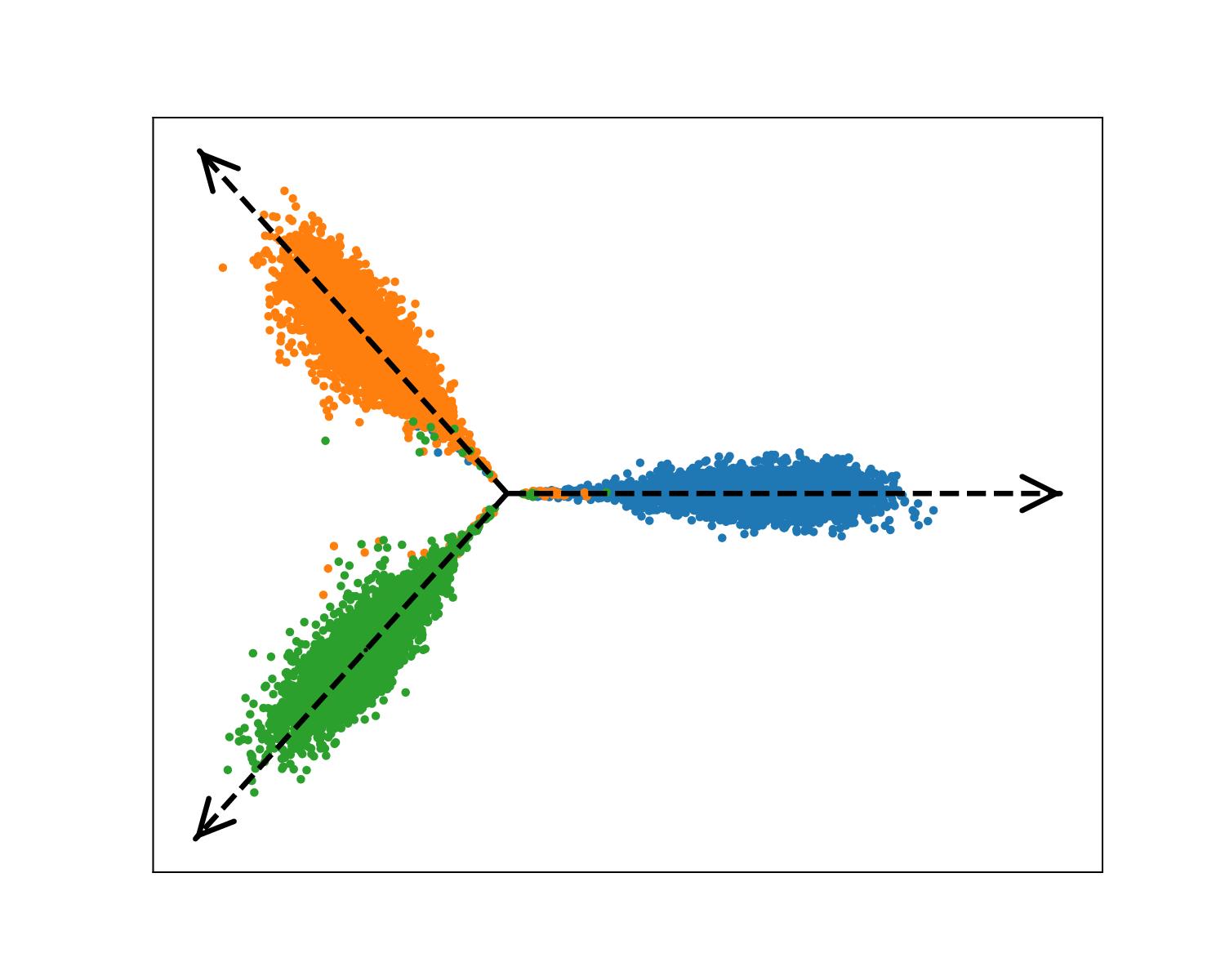}
    \end{subfigure}
    \begin{subfigure}{0.61\textwidth}
        \centering
        \includegraphics[width=0.75\linewidth]{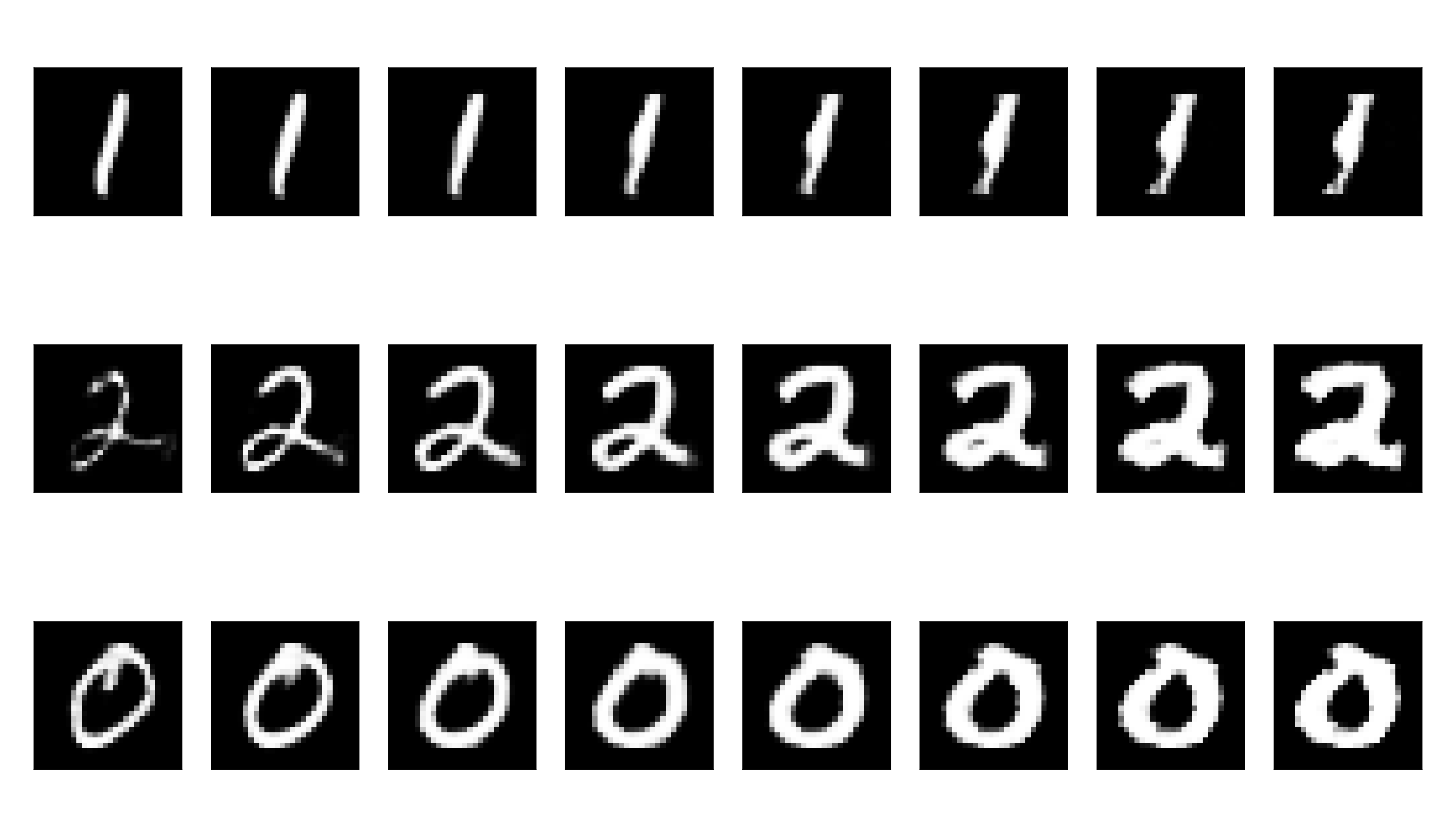}
    \end{subfigure}
    \caption{Generated images along indicated cluster directions of the first two dimensions of the latent space for MNIST012.}
    \label{fig:main_figure}
\end{figure}

In Fig. \ref{fig:main_figure}, the spectral embedding space for MNIST012 shows well-separated clusters where the traversals along the cluster directions give us an indication on which features the model has clustered the data. Compared to the line structure of classical KSC in Fig. \ref{fig:KSC_structure}, the \modelname model has enabled the generation of new points, even beyond the farthest point on the cluster line representing the cluster prototype, allowing us to exaggerate the characteristic feature in the cluster revealing that thinner digits are harder to cluster.

\begin{figure}[h!]
    \centering
    \begin{subfigure}{0.47\columnwidth}
        \centering
        \includegraphics[width=0.77\linewidth]{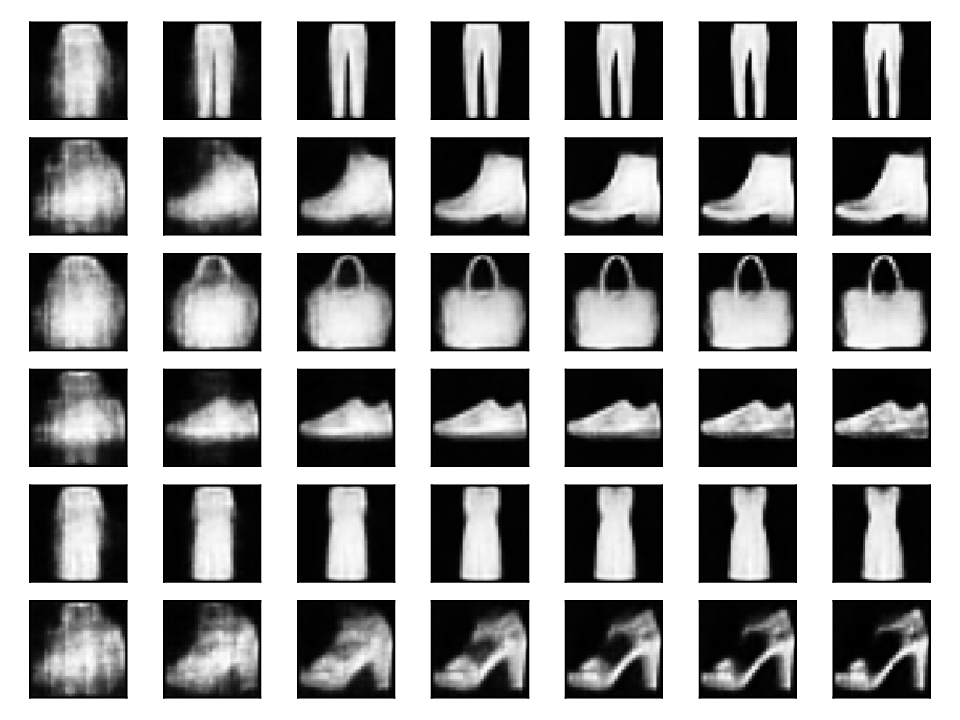}
    \end{subfigure}
    \hfill
    \begin{subfigure}{0.47\columnwidth}
        \centering
        \includegraphics[width=0.77\linewidth]{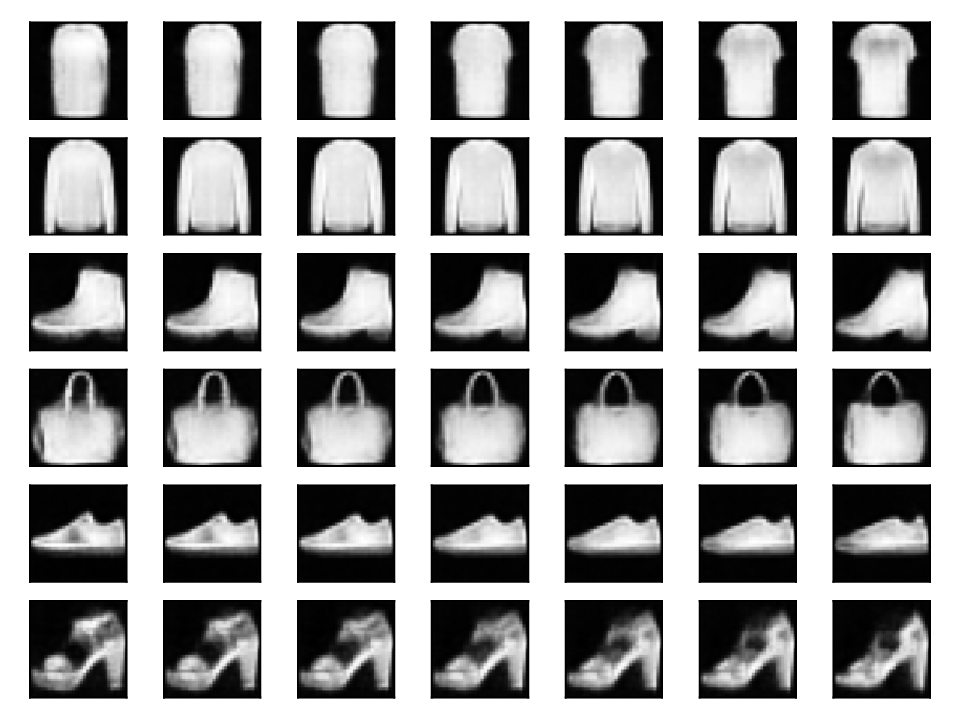}
    \end{subfigure}
    \caption{Latent space traversals for the FashionMNIST dataset. \textbf{Left}: The traversals along the cluster directions in the 9-dimensional latent subspace. \textbf{Right}: Traversals along the $k$-th dimension of the latent space.}
    \label{fig:fashion}
\end{figure}

The model and line structures generalizes to clusters $k > 3$. Fig.~\ref{fig:fashion} shows the traversals along the high-dimensional cluster directions for 6 clusters of the FashionMNIST dataset. Again, the extrapolations in the latent space yield characterizations of the features that are indicative for the clustering. For example, in rows 1 and 5, two pant legs become more distinct, and the shoulder straps of the dress become more prominent. Additionally, generated points along higher components, like shown on the right, enable us to observe intra-cluster variations, such as the distinction between sleeveless and T-shirt sleeves.

\section{Conclusion}
The \modelname model has demonstrated its ability to combine representational learning with a clustering objective to yield an explorable latent clustering space. The key idea is that by extrapolating in the latent space, we generate new data points that emphasize or exaggerate distinctive cluster features---something that existing methods cannot achieve. Future work can include generalizing to a semi-supervised setting and even a fully supervised setting where the cluster labels could be given by another clustering model, potentially creating an interpretable clustering latent space from any clustering model.

\begin{footnotesize}

\bibliographystyle{unsrt}
\bibliography{bibliography}

\end{footnotesize}

\end{document}